\documentclass[final,5p,times,twocolumn]{elsarticle}

\usepackage{amssymb}
\usepackage{longtable}
\usepackage{tabularx}
\usepackage{booktabs}
\usepackage{afterpage}
\usepackage{wrapfig}
\usepackage{soul}
\usepackage{xcolor}
\usepackage{comment}
\usepackage[colorlinks=true, allcolors=blue]{hyperref}

\journal{Energy and Buildings}

\begin{document}

\begin{frontmatter}



\title{\textbf{DECODE: Data-driven Energy Consumption Prediction leveraging Historical Data and Environmental Factors in Buildings}}


\author[label1]{Aditya Mishra\corref{label2}}
\ead{aditya21@iiserb.ac.in}
\cortext[label2]{Corresponding author at: Department of EECS, IISER Bhopal, India } 
\affiliation[label1]{organization={Department of Electrical Engineering and Computer Science},
            addressline={Indian Institute of Science Education and Research (IISER)},
            city={Bhopal},
            postcode={462066},
            state={M.P.},
            country={India}}



\author[label1]{Haroon R. Lone}

\author[label3]{Aayush Mishra}

\affiliation[label3]{organization={Department of Physics},
            addressline={Indian Institute of Science Education and Research (IISER)}, 
            city={Bhopal},
            postcode={462066}, 
            state={M.P.},
            country={India}}

\begin{abstract}
Energy prediction in buildings plays a crucial role in effective energy management. Precise predictions are essential for achieving optimal energy consumption and distribution within the grid. This paper introduces a Long Short-Term Memory (LSTM) model designed to forecast building energy consumption using historical energy data, occupancy patterns, and weather conditions. The LSTM model provides accurate short, medium, and long-term energy predictions for residential and commercial buildings compared to existing prediction models. We compare our LSTM model with established prediction methods, including linear regression, decision trees, and random forest. Encouragingly, the proposed LSTM model emerges as the superior performer across all metrics. It demonstrates exceptional prediction accuracy, boasting the highest $R^{2}$ score of 0.97 and the most favorable mean absolute error (MAE) of 0.007.
An additional advantage of our developed model is its capacity to achieve efficient energy consumption forecasts even when trained on a limited dataset. We address concerns about overfitting (variance) and underfitting (bias) through rigorous training and evaluation on real-world data.  In summary, our research contributes to energy prediction by offering a robust LSTM model that outperforms alternative methods and operates with remarkable efficiency, generalizability, and reliability.
\end{abstract}




\end{frontmatter}


\section{Introduction}
\label{sec:introduction}
Accurate energy prediction holds significant importance in various aspects of energy management, ranging from efficient resource allocation to the creation of resilient energy systems. It plays a vital role in mitigating the negative impacts of energy supply and demand disparities. Notably, buildings represent a substantial portion of energy consumption, utilizing around 30\% of the total energy produced and contributing to approximately 27\% of energy-related emissions\footnote{\url{https://bit.ly/3ElUl0G}}.

Deploying scalable Building Energy Management Systems (BEMS) \cite{zhao2012review} has become increasingly prevalent in buildings. These systems optimize energy consumption patterns within buildings. Energy prediction assumes a crucial role in ensuring the seamless functioning of such BEMS, as accurate forecasts enable these systems to make informed decisions and adjustments. In essence, energy prediction is a cornerstone for achieving sustainable resource utilization, designing resilient energy infrastructures, and driving the effectiveness of BEMS in modern buildings.

Researchers have proposed numerous approaches for predicting energy consumption in buildings. For example, Physics-based models make predictions by considering inputs like the building's physical attributes (such as floor plan and design) and integrate them with external environmental factors like outside temperature and wind speed \cite{khalil2022machine}. However, utilizing such models requires a deep understanding of the domain to capture the interdependencies among these factors accurately and their influence on energy consumption \cite{gonzalez2017data}. Furthermore, such models can also be computationally intensive~\cite{li2014building}.

Machine Learning (ML) models such as linear regression, random forest (RF), support vector regression (SVR), Bayesian approaches, and auto-regressive moving average (ARIMA)  have found great success in predicting energy consumption \cite{wang2018random,zhong2019vector,yuan2017forecasting,nichiforov2017energy}. 
However, studies show that the accuracy of these shallow models decreases in the presence of volatile variables such as weather data, occupancy count, and sudden events (e.g., power outrage, etc.) \cite{wang2019review}. 
This limitation arises from their inherent structure of possessing only one hidden layer or no hidden layer, which results in hand-engineered feature selection, weak generalization capability, and sample complexity \cite{wang2019review}. Sample complexity refers to the number of training instances that ML models require to successfully learn the target variable (i.e., energy consumption). When faced with abundant energy consumption data, these shallow models could encounter network instability and difficulties in parameters (weights and biases) convergence \cite{wang2019review}. 

On the other hand, Deep learning (DL) models frequently surpass traditional ML models due to their ability to autonomously and hierarchically extract features from raw data, negating the need for labor-intensive manual feature engineering. Within DL models, Long Short-Term Memory networks (LSTMs) stand out by comprehending sequential data characteristics and employing these patterns to forecast subsequent outputs. Notably, LSTMs consider both prior and current inputs to compute the output. As a result, LSTMs have exhibited impressive capabilities in the energy forecasting field~\cite{sehovac2020deep}. However, existing LSTM models' efficacy has only been evaluated on residential or commercial buildings~\cite{hamayat2023deep,alsharekh2022improving,jogunola2022cblstm,barzola2022comparisons,ploysuwan2019deep}. The energy consumption patterns of residential and commercial buildings differ, i.e., consumption patterns in commercial buildings are a bit deterministic as such buildings usually have zero consumption during night and weekend hours, and the appliance usage throughout working hours is primarily static. However, energy consumption keeps varying 24x7 in residential buildings due to several factors (e.g., occupancy,  appliance usage). So, while developing an energy prediction algorithm, evaluating its robustness and generalization on both residential and commercial buildings is essential.    
 
In this work, we develop an LSTM model to predict buildings' energy consumption and evaluate the model using publicly available I-Blend~\cite{rashid2019blend}, a dataset containing 52 months of energy consumption data of seven residential and commercial buildings of an Indian academic Institute.  
Furthermore, we compare its performance with state-of-the-art ML models such as linear regression, decision tree (DT), and random forest (RF). Our model takes into account historical energy consumption, environmental factors such as humidity, temperature, dynamic behavior of occupants, and calendar. With rigorous evaluation, we show that our proposed LSTM model outperforms the state-of-the-art ML models.

The proposed LSTM model achieves good performance even on minimal training data, which makes it optimal for short-term, medium-term, and long-term prediction with excellent reliability. Short-term energy consumption prediction entails forecasting energy usage patterns for the near future, which can span from a few minutes to a week. On the other hand, medium-term prediction extends from about a week to a few months, while long-term prediction encompasses energy consumption forecasts over periods exceeding several months or a year \cite{al2005long}. Following are the contributions of our work.
\begin{enumerate}
    \item We evaluated the generalizability of our proposed model on a unique dataset containing both residential and commercial buildings data. 
    \item We also evaluated the model for short, medium, and long-term predictions. 
    \item Furthermore, we provide a comparative analysis of the proposed LSTM model with other traditional ML models.
\end{enumerate}

Section \ref{sec:related_work} discusses the related work. Section \ref{sec:methods} describes our methodology. Section \ref{sec:parameter_seting} explains the  models' parameters and their optimal values. Section \ref{sec:results} explains the results and \ref{sec:discussion} provides the analysis, and finally Section \ref{sec:conlusion} concludes the paper.  

\section{Related Work} \label{sec:related_work}

Energy consumption forecasting in buildings has witnessed a multitude of endeavors, encompassing diverse mathematical, statistical, and machine learning models~\cite{ahmad2020review}. This literature review delves into recent advancements in predictive methodologies, focusing on both machine learning (ML) and deep learning (DL) based approaches. Specifically, here we discuss the most recent and relevant DL prediction methods.

Mahjoub et al.~\cite{mahjoub2022predicting} proposed a novel approach using Long Short-Term Memory (LSTM) models for short-term energy consumption forecasting. Their investigation included a comparative analysis with the Gated Recurrent Unit (GRU) and the Drop-GRU. The outcomes demonstrated that the LSTM model consistently outperformed the GRU and Drop-GRU counterparts. However, it is noteworthy that the predictive accuracy of the LSTM model marginally decreased when forecasting 15 days compared to the 1-day prediction horizon. Specifically, the Mean Absolute Error (MAE) and $R^{2}$ values changed from 0.039 and 0.96 to 0.058 and 0.73, respectively.

Lotfipoor et al.~\cite{lotfipoor2020short} introduced a hybrid methodology that integrates Convolutional Neural Networks (CNN) with LSTM for short-term energy forecasting, focusing on a 5-minute prediction horizon. Their comprehensive evaluation encompassed a comparison against conventional techniques such as ARIMA, Light Gradient Boosting Machine (LightGBM), Random Forest (RF), and Deep Neural Networks (DNN). The study revealed the superiority of the CNN-LSTM hybrid model, showcasing remarkable performance that exceeded all other models. The achieved MAE and Mean Squared Error (MSE) values of 0.18 and 0.15 were particularly noteworthy, respectively. Similarly, Sehovac et al.~\cite{sehovac2019forecasting} explored the  sequence-to-sequence (S2S) architecture, implemented on both the Gated Recurrent Unit (GRU) and LSTM models for energy consumption forecasting of a commercial building. The investigation underscored the preeminence of the GRU S2S configuration across short-term, medium-term, and long-term predictions.

Abraham et al.~\cite{abraham2022predicting} delved into a comprehensive exploration of energy consumption prediction, employing a combination of LSTM and Convolutional Neural Network (CNN) models. Their study harnessed a dataset spanning fourteen years of energy usage. Impressively, the findings indicated that the LSTM model surpassed the CNN model in predictive accuracy, achieving an error rate of 0.45\% compared to CNN's 4.27\%. Besides the mentioned works, Table~\ref{table:related_work} presents an overview of complementary works, encapsulating pertinent features, dataset specifics, utilized model types, and prediction horizons.

\afterpage{
\clearpage
\begin{table*}[h!]
\centering
\caption{Classification of related prediction based research works.}
\vspace{0.2cm}
\label{table:related_work}
\begin{tabular}{cp{4cm}p{4.2cm}p{3.8cm}p{3.5cm}}
\toprule
\textbf{Ref.} & \textbf{Dataset} & \textbf{Features} & \textbf{Model} & \textbf{Prediction} \\
\midrule
\textbf{\cite{fekri2021deep}} & Three years data of five residential buildings  & Five meteorological and  six temporal & Online Adaptive RNN  & Short \& medium-term\\
\hline
\textbf{ \cite{ahmad2017trees}} & 15 months of hotel data &  Occupancy, temperature, temperature, wind speed, and relative humidity & Feed-Forward Back-Propagation ANN and Random Forest & Short \& medium-term\\
\hline
\textbf{\cite{ullah2019short}} & Four years household data & Nine features including global active power, global reactive power, global intensity, voltage, date, time, etc. & CNN and Multilayer Bi-Directional LSTM (MB-LSTM)  & Short-term \\
\hline
\textbf{\cite{agrawal2021week}} & Four years of household data &  Global active power, global reactive power, voltage, global current intensity, and 3 active energy submetering features & 1-D CNN & Short-term \\ 
\hline
\textbf{\cite{ploysuwan2019deep}} & Residential house of 4 occupants & Twenty-eight features from IoT sensors comprising of load lights, temperature, relative humidity, pressure, wind speed, etc. & Deep CNN and LSTM & Short \& medium-term \\ 
\hline
\textbf{\cite{barzola2022comparisons}} & Ten months from academic buildings & Forty-five energy and environmental features from February - November 2021 & Multi-layer Perceptron (MLP), GRU, and LSTM & Short \& medium-term \\
\hline
\textbf{\cite{jogunola2022cblstm}} &  Residential buildings & Temperature, humidity, wind speed, dew point, and week index--weekday, weekend, bank holiday & CNN and Bi-LSTM & Medium \& long-term \\
\hline
\textbf{\cite{alsharekh2022improving}} & UCI repository’s IHEPC and PJM datasets of residential buildings &   Date, time, global active power, global reactive power, global intensity, ampere (A)3, and voltage & R-CNN with ML-LSTM & Short-term\\
\hline
\textbf{\cite{hamayat2023deep}} & Four years   residential building & Date, time, global active power, global reactive power, voltage, global intensity, \& submetered data & CNN and Bi-LSTM & Medium \& long-term\\
\hline
\textbf{\cite{cheng2020powernet}} & One year of academic buildings & Energy , weather and calendar   & Neural Networks & Medium \& long-term \\
\hline
\textbf{\cite{pham2020predicting}} & One year data of residential buildings & Historical energy data & Random Forest & Short-term \\
\hline
\textbf{\cite{fan2019deep}} & One year data of academic building& Date, time, outdoor weather variables, and operating parameters of a chiller plant & Multi-Layer Regression (MLR), ANN, SVR, Extreme Gradient Boosting Trees (XGB)   & Short-term \\
\hline
\textbf{\cite{khan2020electrical}} & IHEPC and AEP residential datasets & IHEPC dataset consist of global active power, global reactive power, global intensity, date, time \& submetered data, and AEP consist of various temperature, pressure, humidity, windspeed, dew point, etc. variables & CNN and MB-GRU   & Short-term \\
\bottomrule
\end{tabular}
\end{table*}
\clearpage
}

{\it Research Gaps:}
We identify the following research gaps with the literature review. 
\begin{enumerate}
    \item The absence of a unified model tailored for forecasting energy consumption in both residential and commercial buildings. Real-world scenarios often involve mixed-use or dual-purpose spaces where residential and commercial areas coexist. A unified model ensures accurate predictions for buildings with diverse functions, facilitating seamless integration. Beyond practicality, developing and maintaining separate models for residential and commercial buildings can be resource-intensive. A unified approach streamlines development processes and reduces the complexity of managing multiple models, contributing to overall resource efficiency.
  \item  Existing energy prediction models excel in specific prediction horizons (e.g., short, medium, or long-term). However, decision-makers, such as building managers and energy planners, often require insights at different time horizons for effective planning and operational efficiency. 
\end{enumerate}

Building on the research gaps, we aim to develop and test a unified prediction model for both commercial and residential buildings. Furthermore, the model should provide a holistic short, medium, and long-term view for decision-making, supporting day-to-day operations as well as strategic planning.

\section{Methodology} \label{sec:methods}
\begin{figure*}[h!]
    \centering
    \includegraphics[width=\textwidth]{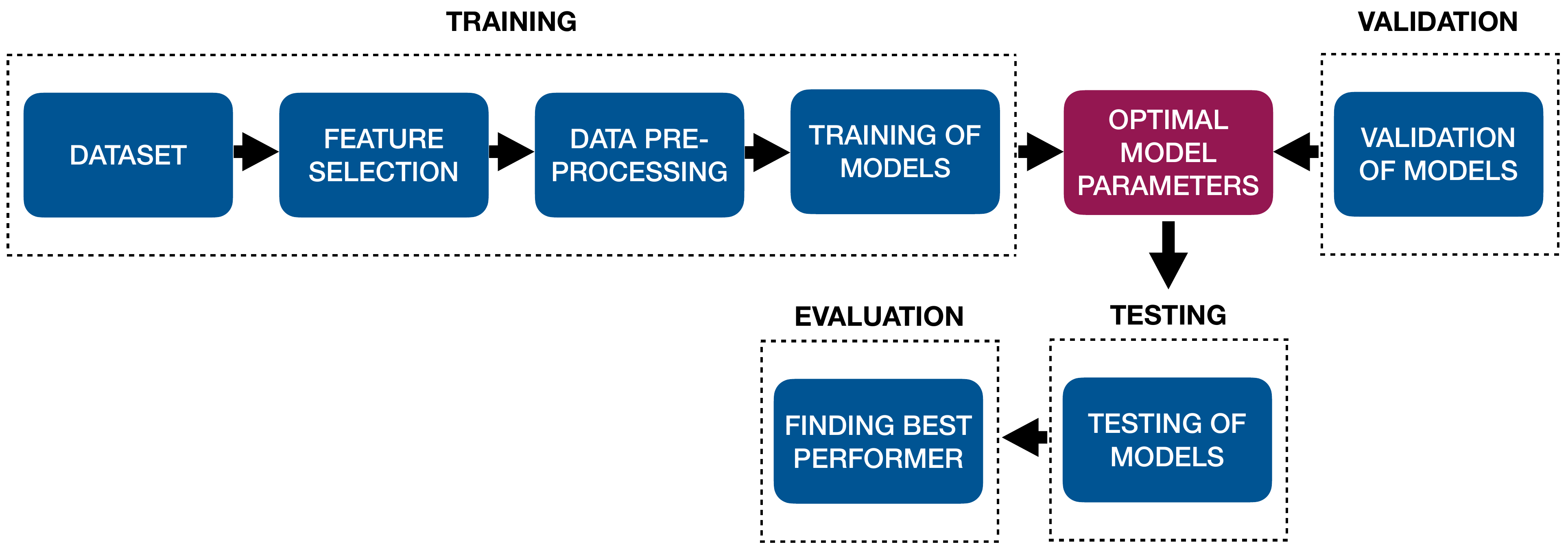}
    \caption{Methodology Flowchart: 
    We initially perform an in-depth analysis of the dataset, involving feature selection and pre-processing to ensure data integrity. Subsequently,  
    the model learns to forecast energy values in the training phase, and the  hyper-parameters are fine-tuned in the validation phase to optimize the model. The models then undergo testing using a separate test dataset, and their  performance is analysed to help us identify the most proficient forecasting model.}
    \label{fig:research_flowchart}
\end{figure*}

We develop and evaluate our methodology on a publicly available I-BLEND dataset~\cite{rashid2019blend}. The dataset contains 52 months of energy data at one-minute sampling rate of seven different residential and commercial buildings of an Indian university. The buildings include academic building (ACB), boys hostel (BH), girls hostel (GH), library (LIB), lecture building (LCB), dining building (DB), and facilities building (FB). Besides energy data, I-BLEND contains occupancy data and environmental factors (temperature, humidity) at sampling rates of 10 and 30 minutes, respectively. Further, it includes an academic calendar showing working and non-working days. Following is a description of our methodology, which is also depicted in Figure \ref{fig:research_flowchart}. 

\noindent{\bf 1. Feature Selection:}
The dataset incorporates additional parameters, including power factor, voltage, and frequency. However, their impact on energy prediction is negligible as these parameters predominantly remain constant throughout the dataset. This observation is substantiated by a feature importance analysis, which consistently ranks these parameters lower in relevance compared to others. As a result we selected occupancy, temperature, calendar, humidity, and historical energy consumption for the prediction purposes.

\noindent{\bf 2. Data Pre-processing:} In this step, we fused energy, occupancy count, temperature, humidity, and calendar data to create a dataset for our analysis. However, the different sampling rates of energy consumption (1 minute), occupancy count (10 minutes), and temperature and humidity (30 minutes) resulted in irregularities.
So, we downsampled energy data and upsampled temperature and humidity data to a common 10 minutes sampling rate by using the merge dataframe of \textit{Pandas} \cite{mckinney2012python}. 
 The missing values for temperature and humidity were interpolated using the \textit{time interpolation} function of \textit{Pandas} \cite{mckinney2011pandas}. \textit{Time interpolation} is a method to search for missing values between the data points by considering the index rather than ignoring it as in the case of \textit{linear interpolation}~\cite{panda2020method}. Moreover, it works on daily and high-resolution data - hourly or minute-level data. Some data points from the girls hostel (GH) building with all the features are shown in Table \ref{table:data}.

 Table~\ref{table:df_scales} shows a high-level description of our dataset. The dataset features are at varying scales highlighting the need of data normalization. The normalization process enhances data integrity, mitigates issues caused by dominant features, and aids in convergence during training. For normalization, we employed the Min-Max normalization, a linear technique that transforms the data feature $X$ to $ X'$ as follows:

 \begin{equation}
   X' = \frac{X - Min(X)}{Max(X)-Min(X)}   
\end{equation}

where $X=[x_{1},x_{2},...,x_{n}]$ is the set of original feature values, $X'=[x'_{1},x'_{2},...,x'_{n}]$ is the set of normalized values of $X$, and $Min(X)$ and $Max(X)$ are the minimum and maximum values of the feature, respectively.

\noindent{\bf 3. Data Division:}
We divided the dataset into training, validation, and testing in the ratio of 70:15:15.
The dataset was divided chronologically, starting from 2013-12-31 at 18:30:00, to ensure that the training, validation, and test sets maintain the temporal order of the data. The forecasting of future energy consumption in buildings relies on historical energy consumption data. Therefore, the dataset has been partitioned in chronological order, considering past energy consumption patterns to enhance the accuracy of predictions for future energy consumption. This chronological splitting strategy also ensures that the model is evaluated on data it has not seen during training, which helps to assess its generalization performance effectively.

\begin{table*}[h!]
\centering
\caption{First five entries of the girls hostel dataset with the features timestamp, energy, occupancy count, temperature, humidity and calendar at a sampling rate of 10 minutes. The datasets for the other buildings exhibit similar entries. Humidity, being a relative measure, does not have units.}
\label{table:data}
\begin{tabular}{ccccccc}
\toprule
\textbf{} & \textbf{Timestamp} & \textbf{Energy (Wh)} & \textbf{Occupancy (\#)} & \textbf{Temperature (\textdegree C)} & \textbf{Humidity} & \textbf{Calendar} \\
\midrule
\textbf{0} & 2014-02-15 18:50 & 4196.86 & 40 & 11 & 100 & 0  \\ 
\textbf{1} & 2014-02-15 19:00 & 4265.65 & 56 & 11 & 100 & 0    \\
\textbf{2} & 2014-02-15 19:10 & 4162.51 & 56 & 11 & 100 & 0 \\
\textbf{3} & 2014-02-15 19:20 & 4730.01 & 60 & 11 & 100 & 0 \\
\textbf{4} & 2014-02-15 19:30 & 4169.19 & 61 & 11 & 100 & 0 \\
\bottomrule
\end{tabular}
\end{table*}

\begin{table*}[h!]
\centering
\caption{Descriptive statistics of the lecture building (LCB). The table underscores the significant variations in numerical range of features illustrating the necessity of data normalization. Similar trends were observed in other buildings.}
\label{table:df_scales}
\begin{tabular}{p{3cm}ccccc}
\toprule
\textbf{} & \textbf{Energy (Wh)} & \textbf{Occupancy (\#)} & \textbf{Temperature (\textdegree C)} & \textbf{Humidity} & \textbf{Calendar} \\
\midrule
\textbf{Count} & 151516 & 151516 & 151516 & 151516 & 151516  \\ 
\textbf{Mean} &  746.6 & 39.9 & 27.2 & 56.7 & 0.7   \\
\textbf{Standard Deviation} & 1588.8 & 71.5 & 7.7 & 23.4 & 0.5 \\
\textbf{Minimum} & 0  & 1 & 2 & 5 & 0\\
\textbf{Maximum} & 26501.1 & 498 & 48 & 100 & 1\\
\bottomrule
\end{tabular}
\end{table*}

\noindent{\bf 5. Prediction Methods:} 
In this section, we describe proposed Long Short-Term Memory (LSTM) model and other machine learning methods for predicting energy consumption. The input variables for the models are historical energy consumption, occupancy count, temperature, humidity, and calendar. The output or the target variable is the predicted energy consumption. We use linear regression (LR), decision tree (DT), random forest regressor (RF), and long short-term memory (LSTM) as these machine learning (ML) models are efficient and have been widely used in forecasting problems \cite{bianco2009electricity, tso2007predicting,wang2018random,kim2019predicting}. Following is a description of the mentioned methods.

\textbf{i. Long Short-Term Memory (LSTM):} It captures long term dependencies in the sequential data and is widely used in time-series forecasting problems. It contains several hyperparameters including {\tt number of LSTM layers, number of units or memory cells in LSTM layers, number of dense layers, number of units in dense layer, batch size, and number of epochs}. Each one of these hyperparameters plays a pivotal role in defining the model's performance. The {\tt number of LSTM layers} and {\tt number of memory cells} help the model to learn and remember long-range dependencies. The role of {\tt dense layers} and {\tt number of dense layer units} is to combine the information from all units or memory cells in the previous layer, allowing the LSTM to capture intricate patterns and dependencies in the data. {\tt Batch size} refers to the number of samples processed before updating the model's weights, and one complete iteration through the entire training dataset is known as one {\tt epoch}. 
LSTM models optimize model parameters (i.e., weights and biases) by minimizing loss function with an {\tt optimization algorithm}. Examples of such optimization algorithms include RMSprop, Adam, ReLU, Tanh, etc. 

\begin{figure*}[h!]
    \centering
    \includegraphics[scale=0.7]{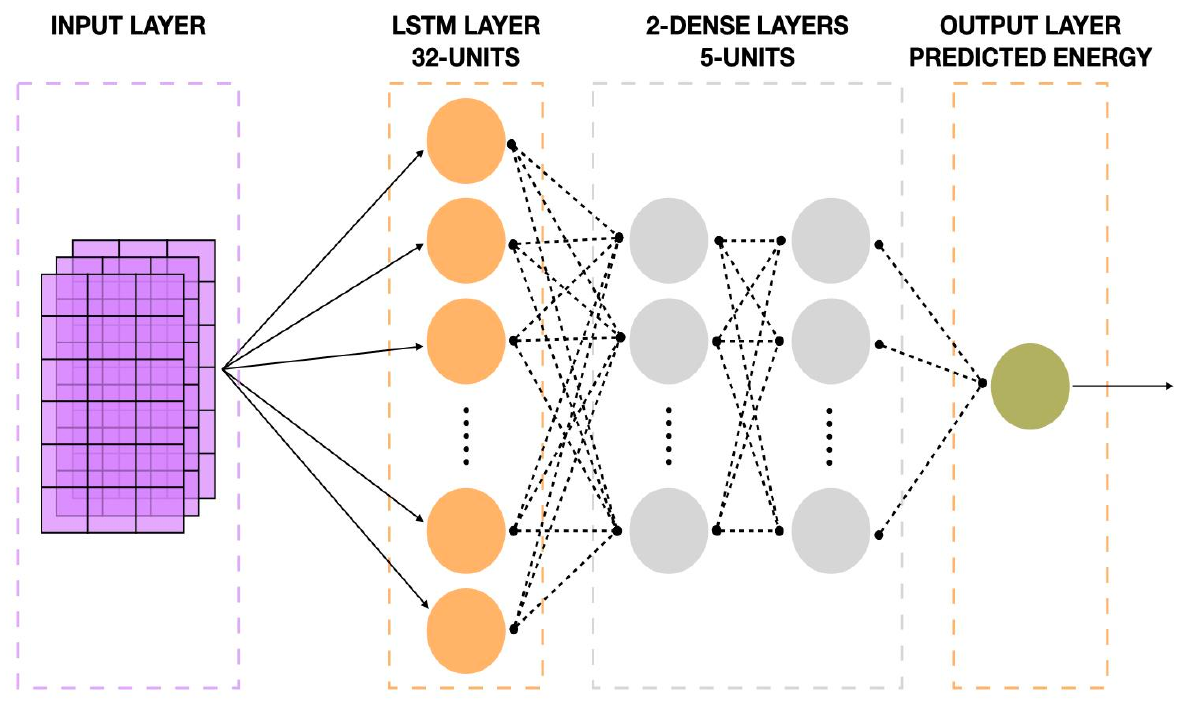}
    \caption{The architecture of proposed LSTM network which is composed of an input sequence, illustrated by purple slabs at the outset of the diagram. It encompasses a single LSTM layer featuring 32-units, denoted by orange dots. Additionally, the model comprises of two dense layers, each containing 5-units, represented by grey dots. Lastly, the output layer, responsible for predicting energy consumption, is depicted by a green dot at the network's end.}
    \label{fig:proposed-lstm-model}
\end{figure*}

{\it Proposed Architecture: }\label{subsec2}
Figure \ref{fig:proposed-lstm-model} shows the architecture of our proposed LSTM model. In the input layer, it takes historical energy consumption, occupancy count, temperature, humidity, and calendar as input; and in the output layer, it predicts energy values. The {\tt number of LSTM units} (or memory cells) is set to 32, determined empirically through tuning and \textit{GridSearch CV}. We performed tuning using \textit{RandomizedSearch CV} to quickly explore and identify a promising region, followed by \textit{GridSearch CV} to perform a more detailed search within that region. These {\tt LSTM units} apprehend and store long-term dependencies in the input sequence by adapting to retain and forget information over time. Two {\tt dense layers}, each containing five units, are added to capture high-level patterns and relations in the sequence. The convergence and performance of the LSTM model depend upon the optimum values of {\tt batch size} and the {\tt number of epochs}, as defined previously.  The {\tt number of epochs} and {\tt batch size} were set to 20 and 64, respectively, determined using \textit{RandomizedSearch CV} followed by \textit{GridSearch CV}. 

We align the input data features for the LSTM model as sequences where each sequence associated with a timestamp comprises of historical energy consumption, occupancy count, temperature, humidity, and calendar values. These input sequences are mapped to each of the 32 units of the LSTM layer.  Each unit captures distinct patterns of the input sequence which are further processed by memory cells of dense layers and finally predict energy. 

We tested the {\tt optimization algorithms} and found that RMSProp achieves the best results. Hence, we use it as the {\tt optimizer} in the LSTM model. It adapts the learning rate for each parameter separately and adjusts the learning rate based on the magnitudes of the recent gradients for each parameter. Moreover, it controls the step sizes taken during the parameter updates, resulting in faster convergence and better performance. 

\textbf{ii. Linear Regression} is a linear approach to predict the energy consumption by using a single known input feature. It works by finding a fit that passes through the maximum number of input data points and minimizes the error, thus increasing the accuracy. Multi-variate linear regression is a type of linear regression that uses multiple input features to predict the energy consumption \cite{maulud2020review}. The parameter {\tt alpha} in the regression model helps to control the extent of regularization. Regularization is the process of adjusting ML models to prevent overfitting or underfitting \cite{tian2022comprehensive}.

\textbf{iii. Decision Tree Regressor} selects an input feature among historical energy consumption, occupancy count, temperature, humidity, and calendar and splits the input dataset into subsets based on a splitting  criterion -  squared error, mean absolute error (MAE), Huber loss, etc. The splitting criterion aims to minimize the variance of the predicted energy consumption within each subset. The criterion chosen for the decision tree model is squared error. Hyperparameters {\tt maximum depth} and {\tt minimum samples split} stop further splitting of nodes and determine the minimum number of samples required to split an internal node, respectively.

\textbf{iv. Random Forest Regressor} provides output by averaging predicted energy consumption of multiple randomized decision trees. It is widely used for classification and regression problems. Moreover, it has shown great success with small as well as large sample sizes and has the advantage of tuning fewer hyperparameters~\cite{biau2016random}. The number of decision tress in the random forest are controlled by the {\tt n-estimators} hyperparameter.

\begin{figure*}[h!]
    \centering
    \includegraphics[width=\textwidth]{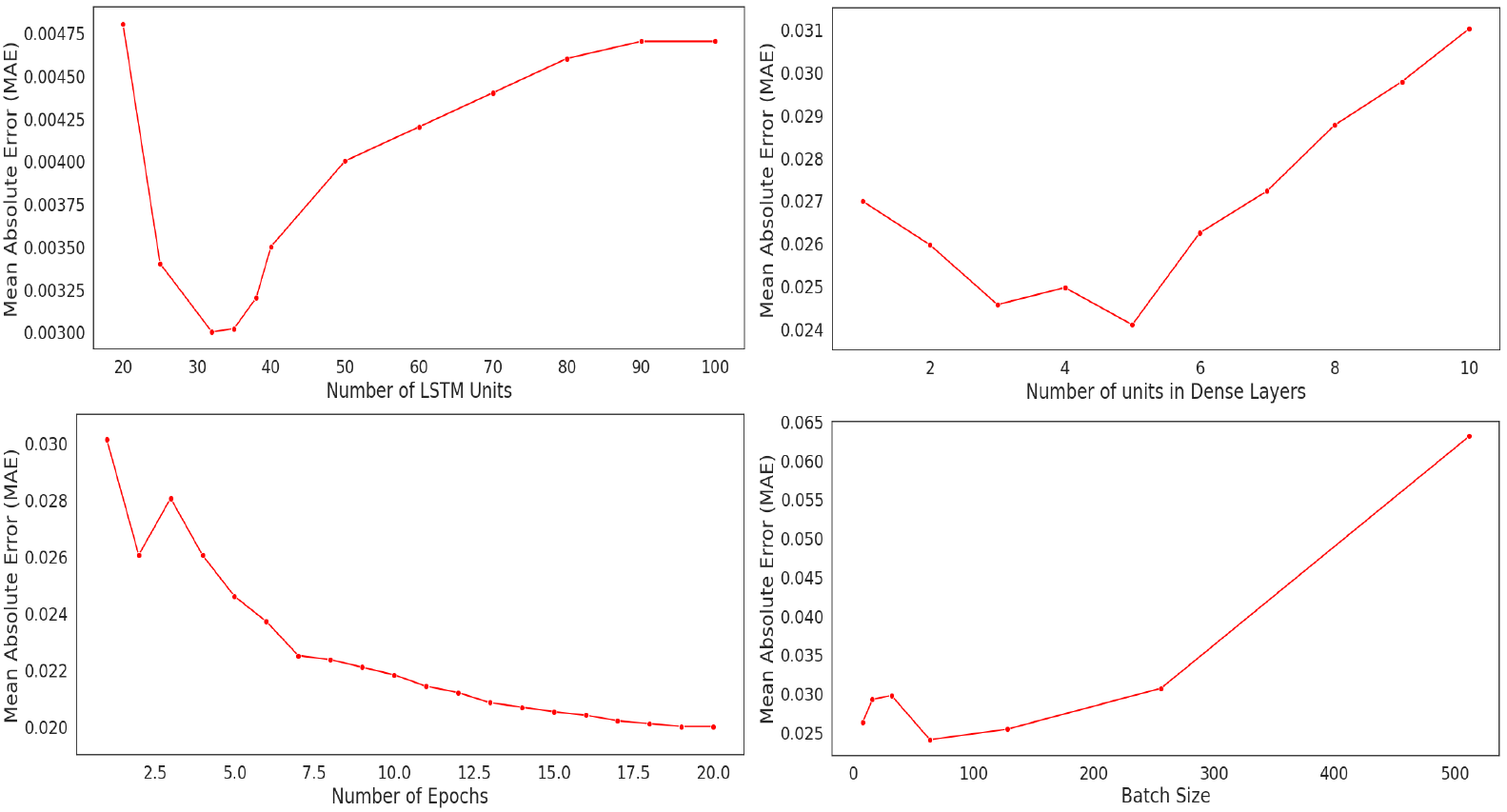}
    \caption{Hyper-parameter tuning: Plots of mean absolute error (MAE) with varying values of hyperparameters - LSTM units, dense layer units, number of epochs, and batch size. The optimum values of hyperparameters were found using \textit{GridSearch CV}. For these plots, we kept all the hyperparameters constant (at the optimum value) and varied one hyper-parameter to obtain the respective variations in MAE.}
    \label{fig:param_search}
\end{figure*}

\section{Parameter Setting} \label{sec:parameter_seting}

In this section we discuss parameter settings of the discussed methods:

{\bf LSTM:} The important hyperparameters in the LSTM model include (i) the {\tt number of LSTM units}, (ii) the {\tt number of units in dense layers}, (iii) the {\tt number of epochs}, and (iv) the {\tt batch size}. We varied the 
 {\tt number of LSTM units} between 20 and 100 as shown in Figure~\ref{fig:param_search}, and found that 32 resulted in lowest mean absolute error (MAE). Similarly, we varied the {\tt number of units in dense layers} between 2 to 10, and found that 2 dense layers comprising of 5 units each were the optimum hyperparameter value.
Also, {\tt batch size} combinations of 8, 16, 32, 64, 128, 256 and 512 were tested, with the {\tt batch size} of 64 yielding the lowest MAE. The optimal combinations  were tested using \textit{RandomizedSearch CV} and \textit{GridSearch CV}. For each iteration, the \textit{RandomizedSearch CV} technique selects a random selection of hyperparameter values and execute a random search over the specified hyperparameter distributions or ranges. Thus, utilizing less computational resources and time. On the contrary, \textit{GridSearch CV} uses an internal cross-validation technique to calculate the score for each hyperparameter combination on the grid. The {\tt number of epochs} was initially set to 5. However, there was no substantial reduction in the loss function after 14 to 15 epochs, hence a maximum of 20 epochs was chosen for the best performance of the model. 

The models - linear regression, decision tree, and random forest, also contain features of the previous three days' energy at a given time instant. For example, the features on 2015-07-05 16:00:00 (Sunday) will contain energy, temperature, humidity, calendar, occupancy count at that time instant, and three days' energy from the previous three non-working days at the same time instant. If it would have been a working day, then it contained three days' energy from the previous three working days. 

{\bf Linear Regression:} Parameter {\tt alpha} characterizes the degree of regularization. We found the optimal value of {\tt alpha} ({\tt alpha} = 1) by varying it in the range of 0.001 to 1000.

{\bf Decision Tree:} Parameter, {\tt maximum depth} stops further splitting of nodes once the maximum threshold is achieved. The optimum value of 14  was found by testing by varying it in the range of 1 to 20. Parameter, {\tt minimum samples split} specifies the minimum number of samples required to split an internal node. It is set to 2 by default. Less value of this parameter results the model to be biased or underfit the training dataset, while a too-large value can result in the detection of differences that are not relevant for the energy consumption prediction or overfit the training data. In either of the conditions, the model's performance degrades. It was tested over the range of 2 to 30, and an optimum value of 20 was observed.

{\bf Random Forest:} Parameter, {\tt n-estimators} refers to the number of tress in the forest. High number of trees enables the model to learn and perform better. However, adding a lot of trees can slow down the training process considerably. Therefore, we did a parameter search to find an optimal value. The optimum value of {\tt n-estimators} was found to be 500 after varying between 10 and 1000.

All the optimum hyperparameter values mentioned above are determined using the \textit{RandomizedSearch CV} and \textit{GridSearch CV} method from \textit{sklearn} library of \textit{Python}.

\subsection{Evaluation metrics}
\label{sec:evaluation_metrics}
We evaluated the performance of the models with two evaluation metrics – $R^{2}$ score, and Mean Absolute Error (MAE).

\noindent{\bf $R^{2}$ score:} The $R^{2}$ score or the coefficient of determination measures the proportion of the variance in the target predicted variable (i.e., energy) explained by the input independent variables (i.e., historical energy consumption, occupancy, temperature, humidity, calendar). 
It varies between 0 and 1.  The higher the score, the better is the prediction model. 

Mathematically, 

\begin{equation}\label{eqn:r2}
  R^{2} = 1 - \frac{\sum_{i=1}^{n}|y_i-\hat{y_i}|^2}{\sum_{i=1}^{n}|y_i-\bar{y}|^2}  
\end{equation}

Where, $\hat{y_{i}}$ and $y_{i}$ refers to the $i^{th}$ predicted and actual values, $\bar{y}$ is the mean of $y$.

\noindent{\bf MAE:} It calculates the average absolute difference between predicted values ($\hat{y}$) and actual values (${y}$). Mathematically, 

\begin{equation} \label{eqn:mae}
    MAE = \frac{\sum_{i=1}^{n}|y_i-\hat{y_i}|}{n}
\end{equation}

Where $n$ refers to the sample size.

MAE is the loss function that need to be minimized in the process of training a machine learning model. Lower MAE values suggest better model performance.

\begin{figure*}[h!]
    \centering
    \includegraphics[scale=0.65]{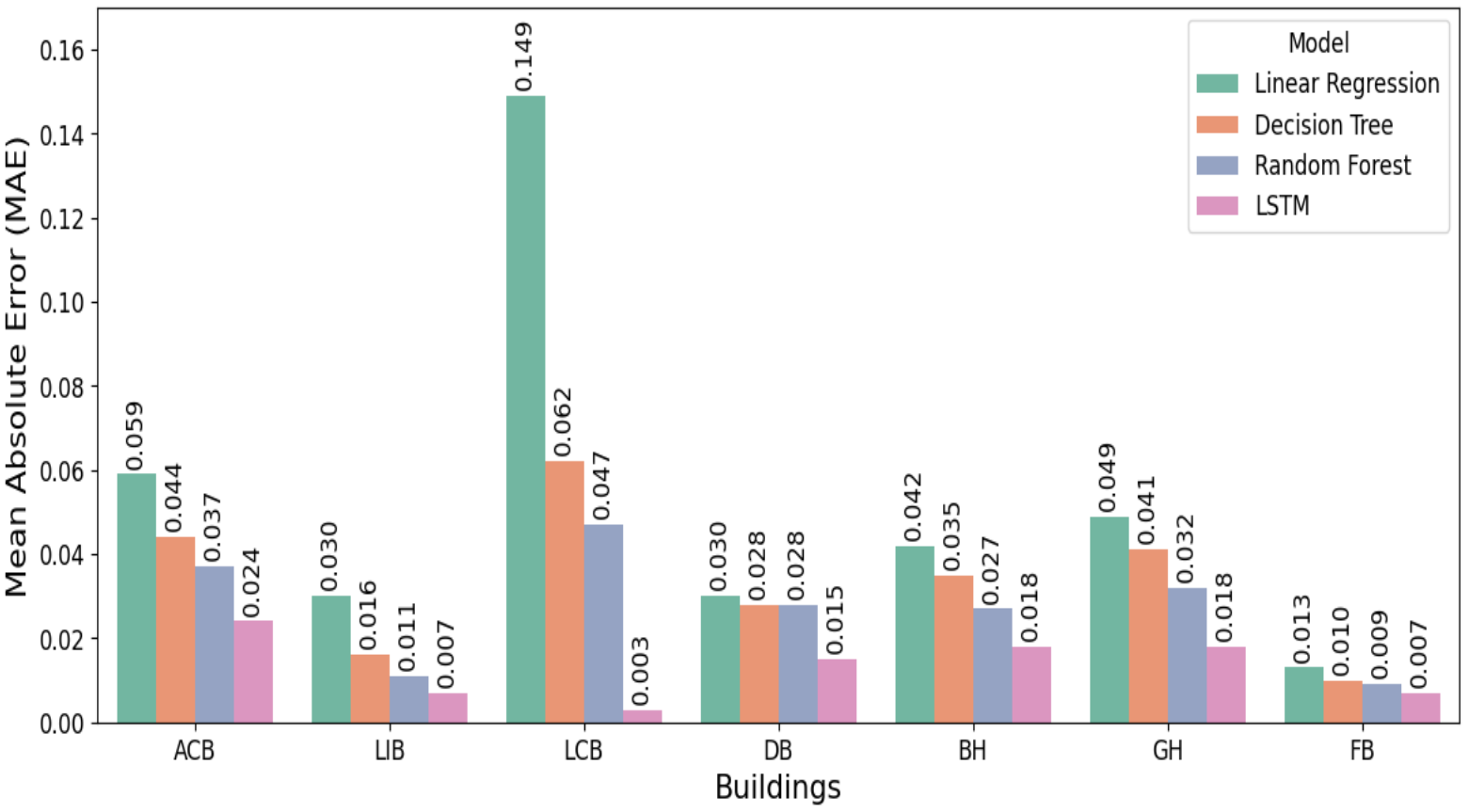}
    \caption{MAE of the proposed LSTM and other ML models in different buildings (ACB - Academic Building, LIB - Library, LCB - Lecture Building, DB - Dining Building, BH - Boys Hostel, GH - Girls Hostel, and FB - Facilities Building). LSTM model performs better as compared to other models as it attains the lowest MAE value on all the buildings.
     }
    \label{fig:mae}
\end{figure*}

\section{Results} \label{sec:results}
Figure~\ref{fig:mae} shows the mean absolute error (MAE) on the test dataset with LSTM, linear regression, decision tree, and random forest. The LSTM model outperforms other models and gives an average MAE value of  0.013. The average $R^{2}$ score of the LSTM model over all the seven buildings is 0.91 (Table~\ref{table:predictions_all}). The ideal $R^{2}$ score shows the efficiency and accuracy of the proposed LSTM model.

\begin{figure*}[p]
    \centering
    \includegraphics[width=\textwidth, height=20cm]{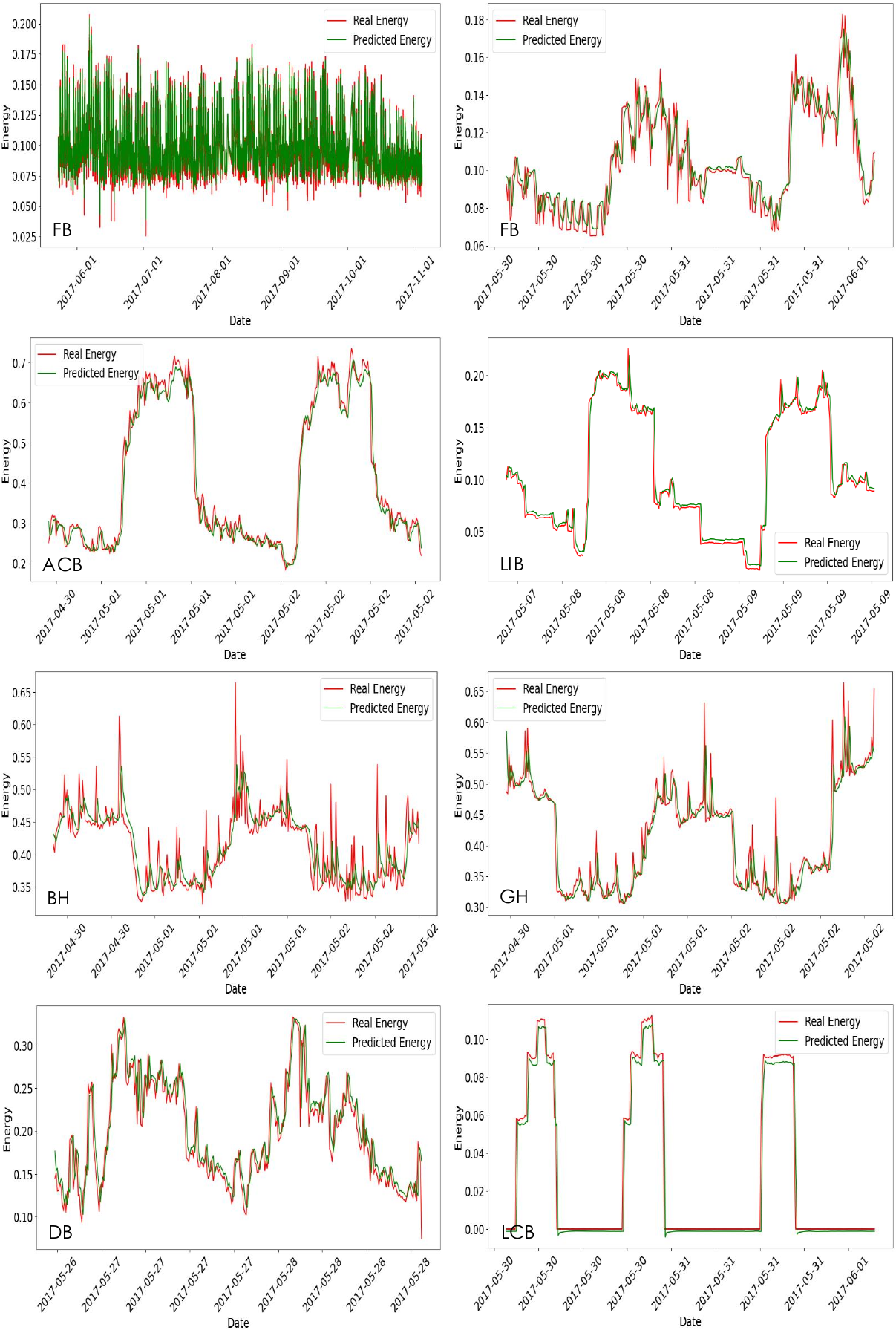}
    \caption{The first plot shows the energy forecast for facilities building (FB) for the entire test dataset. However, the remaining plots show the forecast for facilities building (FB), academic building (ACB), library (LIB), boys hostel (BH), girls hostel (GH), dining building (DB), lecture building (LCB) for initial 50 hours of the test dataset. The accurate predictions of the energy consumption show the efficiency of the proposed LSTM model.}
    \label{fig:prediction}
\end{figure*}

Figure~\ref{fig:prediction} show the energy consumption forecast of all the buildings. The forecast on the entire test data is shown for the facilities building. Since the test dataset is large, thus it is not clearly observable to see the overlapping. For this, we show the forecast plot for the initial 300 timestamps, i.e., 50 hours of the test dataset of respective buildings. The X-axis contains dates, and the Y-axis contains the energy values. The accuracy of our model can be seen as the result of a good overlapping curve of real energy and predicted energy. The red color curve depicts real energy, whereas the green color curve depicts the predicted energy. The dates (X-axis) on each plot vary because the testing was done on 15\% of the length of the respective dataset. 

The proposed LSTM model requires a minimal amount of training dataset for forecasting with a high accuracy, which can be understood by Table \ref{table:minimal_training_data}. It is evident from Table \ref{table:prediction_type} that the developed LSTM model can also be utilized for forecasting short-term, medium-term, and long-term energy consumption.

The efficiency of our LSTM model for the various buildings in our dataset is listed in Table \ref{table:predictions_all}. Our LSTM model performs well in both academic and residential buildings, even with a minimal training dataset, which demonstrates the model's efficiency. It is evident that we require at least 2-3 months of training data for energy forecasting with high accuracy. The ideal $R^{2}$, and lower MAE values are crucial for determining the accuracy of our model.

\begin{table}[h!]
\centering
\caption{$R^{2}$ score and MAE on test data with LSTM model trained on different lengths of training dataset for the academic building. Similar results were obtained for other buildings in our dataset. The $R^{2}$-score ratio, being a dimensionless quantity, is unitless. }
\label{table:minimal_training_data}
\begin{tabular}{p{4.5cm}ccc}
\toprule
\textbf{Length of Training Data} & \textbf{$R^{2}$ - score} & \textbf{MAE (Wh)} \\
\midrule
\textbf{1 Year} & 0.94 & 0.02  \\ 
\textbf{6 Months} &  0.92 & 0.03  \\
\textbf{3 Months} & 0.86 & 0.04 \\
\textbf{2 Months} & 0.83 & 0.05\\
\textbf{1 Month} & 0.53  & 0.08\\
\bottomrule
\end{tabular}
\end{table}

\begin{table}[h!]
\centering
\caption{Performance of the proposed LSTM model on the test data of various time spans for the academic building. The model achieves a very high accuracy for short-term, mid-term, as well as long-term predictions. Similar results were obtained for other buildings in our dataset as well.}
\label{table:prediction_type}
\begin{tabular}{p{2cm}ccc}
\toprule
\textbf{Time Span} & \textbf{$R^{2}$ - score} & \textbf{MAE (Wh)} \\
\midrule
\textbf{1 Day} & 0.96 & 0.028  \\ 
\textbf{1 Week} &  0.96 & 0.023 \\
\textbf{1 Month} & 0.94 & 0.023  \\
\textbf{6 Month} & 0.95 & 0.026 \\
\textbf{1 Year} & 0.96 & 0.024 \\
\bottomrule
\end{tabular}
\end{table}

\begin{table}[h!]
\centering
\caption{$R^{2}$ score and MAE on the test data of all the buildings in our study with the proposed LSTM model. The model acquires good accuracy for all the buildings.}
\label{table:predictions_all}
\begin{tabular}{p{3cm}ccc}
\toprule
\textbf{Buildings} & \textbf{$R^{2}$ - score} & \textbf{MAE (Wh)} \\
\midrule
\textbf{Library} & 0.97 & 0.007 \\ 
\textbf{Academic Building} &  0.96 & 0.024 \\
\textbf{Lecture Building} & 0.96 & 0.003 \\
\textbf{Boys Hostel} & 0.92 & 0.018 \\
\textbf{Girls Hostel} & 0.86& 0.018 \\
\textbf{Facilities Building} & 0.81 & 0.007 \\
\textbf{Dining Building} & 0.89 & 0.015 \\ 
\hline
\textbf{Average} & 0.91 & 0.013 \\
\bottomrule
\end{tabular}
\end{table}

\begin{figure}[h!]
    \centering
    \includegraphics[scale=0.33]{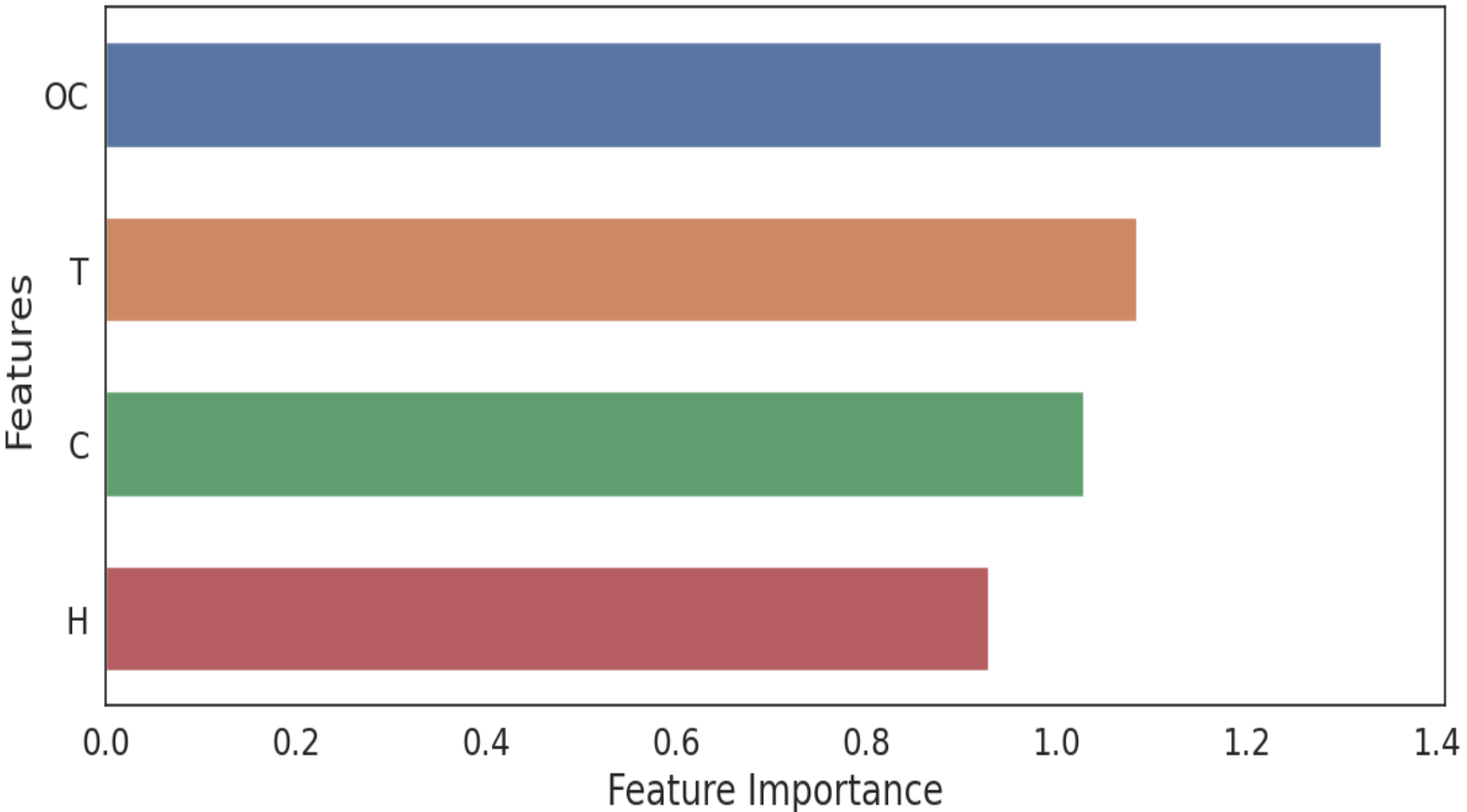}
    \caption{Feature importance showing the relative importance of various features towards prediction in the academic building. In the plot, acronyms OC, T, C and H represent occupancy count, temperature,  calendar, and humidity, respectively.}
    \label{fig:feature_imp}
\end{figure}

\section{Discussion} \label{sec:discussion}

Figure \ref{fig:param_search} demonstrates the need of the hyper-parameter tuning in the LSTM model. The Figure shows that the initial increase in the number of LSTM units corresponds to a reduction in MAE, indicative of increasing performance of the model. However, it eventually flattens, signaling diminishing returns, and then ascends due to overfitting from excessive number of LSTM units. Overfitting refers to the condition in which the model becomes too specific to the training data and doesn't work well with new data.

As the batch size increases, there is an initial rise in the MAE. This can be attributed to the model's heightened sensitivity to noisy updates during the initial phase of training. In other words, when the batch size is small, individual sample variations have a more pronounced impact on weight updates, thereby impeding stable convergence. However, as the batch size continues to grow, the MAE gradually decreases to reach an optimal value. This decline is the outcome of more consistent and smoother weight adjustments during training, facilitating faster convergence and, consequently, a reduction in MAE. Nevertheless, caution should be exercised when increasing the batch size beyond its optimal value, typically found around 64. Beyond this point, there is a risk of overshooting the ideal weight parameters, which can lead the model to converge towards a suboptimal solution. As the batch size surpasses its optimum, the model's ability to generalize effectively diminishes, resulting in reduced overall performance and a subsequent increase in MAE.

While increasing the number of epochs, we observe a initial rapid decline in the MAE. As the number of epoch continues to increase, the rate of decline in MAE usually slows down, indicating that that the model is learning more detailed patterns in the data. The curve gets smoother and converges, showing that the model is getting better. However, there is a risk of overfitting if the number of epochs are constantly increasing beyond optimal value of 20. Similar to the LSTM units, as the number of units in dense layer increases, we observe MAE reduces. Increasing the units beyond the optimal value of 5 results an increase in the MAE because of overfitting. Thus, it highlights the delicate balance required in selecting an appropriate values of hyperparameters to ensure optimal model performance.

The dynamic behavior of occupants poses a significant challenge in the field of energy forecasting~\cite{ullah2019short}. In response to this challenge, the proposed LSTM model considers both the occupants' behavior and weather data to predict energy consumption, as illustrated in Figure \ref{fig:feature_imp}. This figure demonstrates the feature importance of various features, including occupancy count, temperature, calendar, and humidity, in forecasting future energy consumption trends. Features with higher scores contribute more to the prediction of the target variable. We employed \textit{SelectKBest} \cite{desyani2020feature} module of the \textit{sklearn} library to compute the feature importance. It works by selecting the top ``k'' most relevant features (among occupancy count, temperature, humidity, and calendar) from the dataset based on their correlation with the target variable (energy consumption). The figure shows that all four features play a role in predicting energy, with occupancy count being the most influential and humidity being the least influential among the selected features. Note that similar patterns, as shown in Figure~\ref{fig:feature_imp}, are consistently observed across various buildings, except for dining and facilities buildings, where distinctive results emerge. Notably, temperature assumes a more crucial role in influencing energy consumption compared to occupancy count in these specific buildings.

\begin{figure}[h!]
    \centering
    \includegraphics[width=\columnwidth]{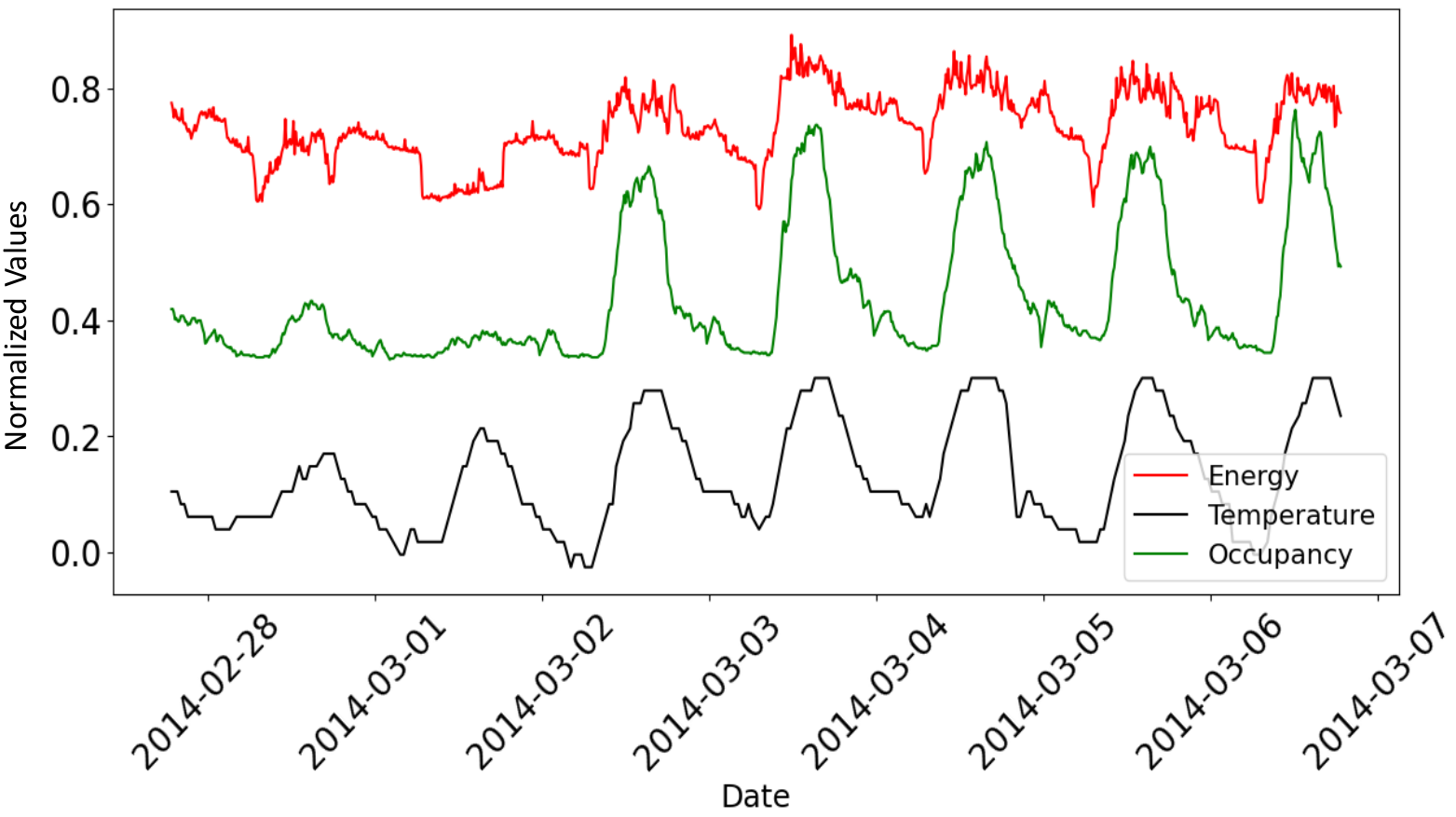}
    \caption{Energy, occupancy count, and temperature of the academic building over one-week. The red line depicts energy, green line shows occupancy count, and the black line represents temperature. The vertical axis represents the normalized values of energy consumption, occupancy count, and temperature.}
    \label{trend1}
\end{figure}

Figure~\ref{trend1} further elucidates these relationships. It illustrates that an increase in the number of occupants leads to a corresponding rise in energy consumption. This can be attributed to the increased usage of appliances such as lights, fans, and charger points. Conversely, a decrease in the number of occupants results in a reduction in energy consumption. Likewise, the figure highlights the influence of temperature on energy consumption. As the day progresses and temperatures rise, energy consumption also increases. This phenomenon results from heightened usage of air-conditioners and increased fan speed to combat the rising temperatures. Conversely, when temperatures decrease, energy consumption tends to decrease as well.
Furthermore, the figure reveals an intriguing connection with the calendar. It shows that on working days, there is a decrease in energy consumption within hostels, while academic buildings and lecture halls experience an increase. This trend is reversed on non-working days, indicating a clear correlation between the type of day and energy consumption patterns.
In summary, this research underscores the intricate interplay between occupants' behavior and environmental factors, shedding light on their collective impact on energy consumption forecasting.

\begin{table}[]
\centering
\caption{Correlation values between energy consumption of different buildings with other features, i.e., occupancy (OC), temperature (T), humidity (H), calendar (C). Bold values represent highest observed correlation for a building.}
\label{table:correlation}
\begin{tabular}{ccccccc}
\toprule
\textbf{} & \textbf{OC} & \textbf{T} & \textbf{H} & \textbf{C} \\
\midrule
\textbf{Library} & \textbf{0.32} & 0.19 & -0.13 & 0.22 \\ 
\textbf{Academic Building} & \textbf{0.66} & 0.43 & -0.25 & 0.33 \\
\textbf{Lecture Building} & \textbf{0.53} & 0.18 & -0.05 & 0.19 \\
\textbf{Boys Hostel} & \textbf{0.52 }& 0.08 & 0.13 & 0.09 \\
\textbf{Girls Hostel} & \textbf{0.46} & 0.25 & 0.03 & 0.12 \\
\textbf{Facilities Building} & 0.26 & \textbf{0.48} & -0.20 & 0.11 \\
\textbf{Dining Building} & 0.36 & \textbf{0.38} & -0.20 & 0.16 \\
\bottomrule
\end{tabular}
\end{table}

The diverse performance of the LSTM model reported in Table~\ref{table:predictions_all} across different buildings can be attributed to variations in input data. Table~\ref{table:correlation} highlights distinct correlations between energy and different features (occupancy count (OC), temperature (T), humidity (H), and calendar (C)) for all the buildings. For all the buildings except facilities and dining, occupancy shows highest correlations with the energy consumption. Facilities building host several HVAC chillers and transformers of the academic Institute and is controlled by a dedicated team of 4 - 5 operators and hence is not affected by the occupancy. Similarly, the dining building is affected by the occupants only during lunch and dinner hours. Furthermore, each building exhibits unique trends in energy consumption across different days. Buildings, where the model successfully captures and learns these trends tend to yield high-performance scores. These include library, academic building, lecture building, and boys hostel. Conversely, in buildings characterized by the absence of clear trends or randomness, the model struggles to learn effectively, resulting in comparatively lower performance scores. These include facilities building, girls hostel, and dining building.

The challenge posed by insufficient training data can lead to a degradation in the model's performance, preventing it from effectively capturing intricate patterns necessary for accurate energy prediction. The outcomes illustrated in Table \ref{table:minimal_training_data} offer a compelling counterpoint. They reveal the commendable performance of our proposed model, even when confronted with a limited volume of training data. This is substantiated by a good $R^{2}$ score of 0.83, and less error value of 0.05. These findings highlight the model's robustness and its ability to extract valuable insights from a modest training dataset.

Most existing works demonstrate competence in either short-term or medium-term energy prediction. However, the distinctive strength of our LSTM model emerges as it excels not only in short-term and medium-term predictions but also in long-term energy consumption forecasting. This robust performance across varying prediction horizons emphasizes the versatility and effectiveness of the proposed LSTM model. The findings presented in Table \ref{table:prediction_type} summarize the conclusive evidence supporting this assertion.

The proposed LSTM model demonstrates proficient energy consumption predictions; however, it exhibits comparatively poor performance when used to forecast energy consumption of the facilities building (FB) and girls hostel (GH). This disparity in the performance can be attributed to the model's challenge in comprehending the trends and patterns between energy consumption and other relevant features specific to the facilities building and girls hostel.

\section{Conclusion \& Future work} \label{sec:conlusion}

This paper proposed an LSTM model for energy prediction that considers the temporal dependencies and includes the dynamic behavior of the occupants and weather data. The model achieved higher accuracy than classical ML models for all seven buildings. The proposed model requires a minimal amount of training dataset, tackles variance and bias problems, and can be used for short, medium, and long-term forecasting. The model's excellent performance on the real-world data makes it reliable.

The model exhibits limitations in cases where anticipated trends between input data, particularly occupancy count and temperature, are not consistently observed within buildings. In such instances, the model may initially struggle to perfectly predict energy consumption. However, it dynamically adapts over time, learning from deviations and adjusting its predictions accordingly.

The future work will include developing an online model using continual learning techniques capable of forecasting based on newly arriving data and adapting to new patterns. Furthermore, online hyper-parameter tuning is required to solve the model's degrading performance issue. In addition to the ongoing development of the online model, future work will extend to explore the realm of model quantization. Model quantization involves compressing the size of the neural network, making it more lightweight and efficient for deployment on resource-constrained devices. Specifically, our research will delve into the implementation of small circuit embedding and integration with embedded systems. This extension aims to enhance the practical applicability of the forecasting model, ensuring its seamless integration into real-world scenarios where computational resources are limited.



\bibliographystyle{elsarticle-num} 
\bibliography{ref}





\end{document}